\documentclass[conference]{IEEEtran}
\IEEEoverridecommandlockouts

\usepackage{cite}
\usepackage{amsmath,amssymb,amsfonts}
\usepackage{algorithmic}
\usepackage{graphicx}
\usepackage{textcomp}
\usepackage{xcolor}
\usepackage{booktabs}
\usepackage{siunitx}
\usepackage{url}

\def\BibTeX{{\rm B\kern-.05em{\sc i\kern-.025em b}\kern-.08em
    T\kern-.1667em\lower.7ex\hbox{E}\kern-.125emX}}

\begin{document}

\title{Benchmarking the Energy Cost of Assurance in Neuromorphic Edge Robotics%
\thanks{Accepted and presented at the STEAR 2026 Workshop on Sustainable and Trustworthy Edge AI for Robotics, HiPEAC 2026, Krakow, Poland.}}

\author{\IEEEauthorblockN{Sylvester Kaczmarek}
\IEEEauthorblockA{\textit{Department of Computing} \\
\textit{Imperial College London}\\
London, United Kingdom \\
research@sylvesterkaczmarek.com}
}

\maketitle

\begin{abstract}
Deploying trustworthy artificial intelligence on edge robotics imposes a difficult trade-off between high-assurance robustness and energy sustainability. Traditional defense mechanisms against adversarial attacks typically incur significant computational overhead, threatening the viability of power-constrained platforms in environments such as cislunar space. This paper quantifies the energy cost of assurance in event-driven neuromorphic systems. We benchmark the Hierarchical Temporal Defense (HTD) framework on the BrainChip Akida AKD1000 processor against a suite of adversarial temporal attacks. We demonstrate that unlike traditional deep learning defenses which often degrade efficiency significantly with increased robustness, the event-driven nature of the proposed architecture achieves a superior trade-off. The system reduces gradient-based adversarial success rates from 82.1\% to 18.7\% and temporal jitter success rates from 75.8\% to 25.1\%, while maintaining an energy consumption of approximately 45 microjoules per inference. We report a counter-intuitive reduction in dynamic power consumption in the fully defended configuration, attributed to volatility-gated plasticity mechanisms that induce higher network sparsity. These results provide empirical evidence that neuromorphic sparsity enables sustainable and high-assurance edge autonomy.
\end{abstract}


\section{Introduction}
Autonomous robotic systems operating in remote and hostile environments, such as cislunar space, face a convergence of severe operational constraints.\cite{b7,b6} These systems must operate with power budgets often below 10 watts for computing, endure high radiation fluxes that induce hardware faults, and function with significant communication delays that preclude real-time human intervention.\cite{b7} Simultaneously, the strategic nature of these domains introduces the risk of adversarial interference, ranging from sensor spoofing to sophisticated data injection attacks.\cite{b8,b20}

Ensuring the reliability of these systems requires the integration of robust defense mechanisms. In conventional deep learning architectures, adding assurance capabilities typically imposes a measurable energy tax.\cite{b16,b21} Techniques such as adversarial training, input purification, or ensemble verification increase the number of floating-point operations per inference, thereby increasing latency and power consumption.\cite{b8,b20} For energy-critical edge robotics, this overhead often renders high-assurance models impractical.

Neuromorphic computing offers a potential solution to this efficiency bottleneck through event-driven processing.\cite{b22} Unlike synchronous von Neumann architectures that consume power based on clock cycles, neuromorphic processors consume dynamic power primarily when processing discrete spike events.\cite{b4,b19} This architectural difference suggests that the energy cost of security might scale differently in spiking neural networks (SNNs) compared to traditional artificial neural networks (ANNs).\cite{b22}

This paper investigates the relationship between adversarial robustness and energy efficiency on commercial neuromorphic hardware.\cite{b5} We define a benchmarking protocol to measure the energy cost of assurance on the BrainChip Akida AKD1000.\cite{b19} We perform a systematic ablation study of the Hierarchical Temporal Defense (HTD) framework, mapping the trade-off between robustness against temporal jitter and gradient-based attacks versus the energy consumed per inference.\cite{b1,b10,b12}

Our primary contribution is the empirical demonstration that specific neuromorphic defense mechanisms can enhance security without imposing an energy penalty.\cite{b1,b5} We show that mechanisms designed to suppress adversarial volatility can inadvertently increase network sparsity, leading to a reduction in dynamic power consumption.\cite{b15} This finding suggests that security and sustainability can be synergistic objectives in the design of neuromorphic edge systems.\cite{b21}

\section{Related Work}
The deployment of machine learning on edge devices is constrained by the size, weight, and power (SWaP) limitations of the host platform.\cite{b16} Research in edge AI has focused heavily on model compression techniques, such as quantization and pruning, to fit deep learning models within these envelopes.\cite{b16} While these methods reduce the computational load, they often do not address the specific security vulnerabilities of the model or the energy cost of adding defensive layers.\cite{b21}

\subsection{Neuromorphic Efficiency}
Neuromorphic computing has matured as a viable alternative for ultra-low-power edge intelligence.\cite{b22} Platforms such as Intel Loihi \cite{b4} and BrainChip Akida \cite{b19,b3} leverage the sparsity of spiking neural networks to achieve orders-of-magnitude improvements in energy efficiency for specific workloads compared to graphics processing units (GPUs).\cite{b22,b17} Prior work has benchmarked the efficiency of these platforms for standard classification and control tasks.\cite{b5,b3} However, these benchmarks typically assume a benign operational environment and do not account for the computational overhead required to secure the model against active threats.\cite{b5}

\subsection{Adversarial Robustness at the Edge}
Adversarial robustness in spiking neural networks is an emerging field.\cite{b9,b10,b11,b12,b13} Recent studies have demonstrated that spiking networks are vulnerable to gradient-based attacks \cite{b20,b8,b10,b11} and temporal perturbations.\cite{b12,b9} Proposed defenses include input filtering, stochastic thresholds, and robust training protocols.\cite{b9,b10,b11,b12,b13} While these studies report improvements in robustness metrics such as attack success rate (ASR), they rarely quantify the impact of these defenses on the hardware-level energy consumption.\cite{b5} This paper bridges this gap by providing a simultaneous evaluation of security and energy efficiency on physical neuromorphic hardware.\cite{b5,b21}

\subsection{Benchmarking Methodologies}
Standardized benchmarking is critical for comparing edge AI solutions.\cite{b16} Benchmarks like MLPerf Tiny focus on latency and energy for standard models but lack specific tracks for adversarial robustness.\cite{b16} Conversely, robustness benchmarks like RobustBench focus on accuracy under attack but do not account for the energy cost of the defense.\cite{b21} Our work aligns with recent calls for holistic benchmarking that treats efficiency and trustworthiness as coupled objectives in the design space of safety-critical systems.\cite{b21,b5}

\section{System Under Test}
This section defines the hardware platform and the assurance architecture used to construct the benchmark. We treat the defense mechanisms as modular configuration layers to isolate their individual contributions to energy consumption and robustness.

\subsection{Neuromorphic Hardware Platform}
The benchmarking target is the BrainChip Akida AKD1000, a commercial neuromorphic system-on-chip (NSoC) designed for edge applications.\cite{b19,b3} The processor is fabricated on a 28 nanometer CMOS process and features a mesh of 80 Neural Processing Units (NPUs).\cite{b19} Unlike conventional accelerators that rely on large matrix multiplications, the Akida architecture is event-driven.\cite{b19,b3} Computation and memory access occur only when a neuron receives a spike event. Consequently, the dynamic power consumption is proportional to the synaptic activity within the network rather than the clock frequency.\cite{b19,b5}

The hardware supports configurable bit-precision for weights and activations, typically ranging from 1 to 4 bits.\cite{b19} For this study, we utilize 4-bit weights to balance model capacity with memory footprint. The chip includes native support for on-chip learning via Spike-Timing-Dependent Plasticity (STDP), allowing synaptic weights to be updated locally on the device without requiring data transfer to a host processor.\cite{b19,b4} This capability is essential for evaluating the energy cost of adaptive defense mechanisms that operate during inference.\cite{b1}

\subsection{Assurance Architecture}
The system under test is a spiking neural network implementing the Hierarchical Temporal Defense (HTD) framework \cite{b1}. This framework provides a multi-layered security architecture designed to mitigate adversarial perturbations in the temporal domain.\cite{b1} For the purpose of this benchmark, we define four cumulative configurations to map the efficiency-robustness trade-off space.

\begin{figure*}[htbp]
    \centering
    \includegraphics[width=0.8\linewidth]{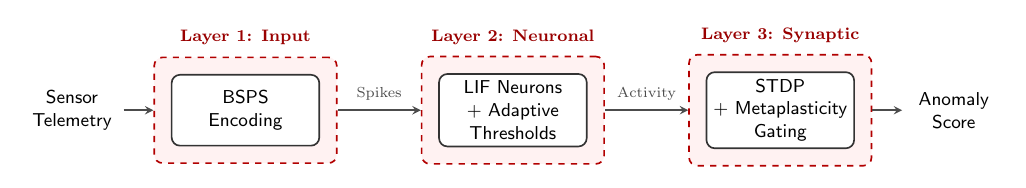}
    \caption{The Hierarchical Temporal Defense (HTD) Architecture used for benchmarking. The system integrates three layers of assurance: (1) Input Assurance via Bayesian Spike Pattern Superposition (BSPS), (2) Neuronal Assurance via Homeostatic Adaptive Thresholds, and (3) Synaptic Assurance via Volatility-Gated Metaplasticity. Each layer filters specific temporal adversarial artifacts before they propagate.}
    \label{fig:architecture}
\end{figure*}

\subsubsection{Baseline Configuration}
The baseline model is a standard spiking neural network without specific defense mechanisms. It utilizes rate coding for input conversion, fixed firing thresholds for Leaky Integrate-and-Fire (LIF) neurons, and standard STDP for online adaptation.\cite{b19} This configuration serves as the reference point for energy and latency measurements.

\subsubsection{Input Assurance Layer}
The first defense layer replaces deterministic rate coding with Bayesian Spike Pattern Superposition (BSPS).\cite{b1} Standard rate coding maps input intensity directly to spike frequency. This method is brittle to temporal jitter where spike times are shifted.\cite{b12} The proposed encoding represents the input signal as a probability distribution over possible temporal patterns within a sliding window $W$.

We encode each input window of length $W$ as a belief state over a finite set of spike pattern hypotheses $\{s_i\}_{i=1}^K$. Given observation $o_t$, the belief update is:
\begin{equation}
p_i(t+1) = (1-\eta)p_i(t) + \eta \frac{P(o_t \mid s_i)p_i(t)}{\sum_{j=1}^K P(o_t \mid s_j)p_j(t)}
\end{equation}
This probabilistic superposition reduces sensitivity to small timing shifts because perturbations produce bounded changes in the belief state rather than discrete symbol flips.\cite{b9}

\subsubsection{Neuronal Assurance Layer}
The second layer introduces Homeostatic Adaptive Thresholds.\cite{b1} In this configuration, the firing threshold of each neuron is dynamic. The threshold increases immediately following a spike event and decays exponentially back to a baseline value. This mechanism acts as a temporal filter, suppressing high-frequency burst patterns often associated with adversarial injections while maintaining sensitivity to nominal signal dynamics.\cite{b12,b9}

\subsubsection{Synaptic Assurance Layer}
The final configuration enables Volatility-Gated Metaplasticity.\cite{b1,b15} This mechanism modulates the learning rate of individual synapses based on the variance of their recent weight updates. Synapses exhibiting high volatility, which is characteristic of gradient-based adversarial attacks, have their plasticity suppressed.\cite{b10,b11} This layer is intended to prevent the corruption of the model during online learning.\cite{b1}

\section{Benchmarking Protocol}
We define a rigorous measurement protocol to quantify the operational cost of the assurance layers described above. The protocol captures detection performance, adversarial robustness, and hardware-level energy consumption.

\subsection{Dataset and Workload}
The workload is derived from the Cislunar Anomaly and Risk Dataset (CARD) \cite{b2}. This dataset contains multi-modal sensor telemetry, including inertial measurement unit (IMU) data and visual feeds, simulating the operational environment of a lunar rover.\cite{b7} The data includes nominal operations and a taxonomy of anomalies ranging from mechanical failures to sensor degradation.\cite{b2,b18} The spiking neural network is tasked with real-time anomaly detection, processing the streaming sensor data to classify system health.\cite{b18}

\subsection{Threat Model}
To evaluate robustness, we subject the system to two distinct classes of white-box adversarial attacks:

\begin{itemize}
    \item \textbf{Projected Gradient Descent (PGD):} This attack iteratively perturbs the input current to maximize the loss function.\cite{b8,b20} It represents a sophisticated gradient-based adversary attempting to force a misclassification.\cite{b8} For SNNs, gradient computation follows standard surrogate-gradient practice.\cite{b14}
    \item \textbf{Temporal Jitter:} This attack adds stochastic delays to the timing of input spikes.\cite{b12} It targets the specific temporal processing mechanisms of the spiking network, attempting to disrupt coincidence detection.\cite{b12}
\end{itemize}

We define the Adversarial Success Rate (ASR) as the percentage of attack attempts that successfully cause the model to misclassify an anomaly as a nominal state.\cite{b8,b12}

\subsection{Measurement Setup}
The experimental setup consists of a host workstation (Intel Core i7) orchestrating the data flow to the Akida AKD1000 via a PCIe interface. Energy consumption is measured empirically using the integrated power monitoring rails on the development board.\cite{b19} We query the power sensors at a frequency of 1 kilohertz during the inference workload.\cite{b19} The total energy per inference is calculated by integrating the instantaneous power draw over the duration of the inference window and dividing by the number of samples processed.\cite{b5}

Latency is measured as the wall-clock time required for the hardware to process a single input window and output a classification event. This includes the time for spike generation and on-chip propagation but excludes the PCIe transfer overhead to isolate the performance of the neuromorphic processor.\cite{b19}

To ensure statistical reliability, all measurements are averaged over 10 independent experimental runs. We report the mean values and standard deviations for energy, latency, and adversarial success rate.\cite{b5}

\subsection{Reproducibility Details}
To facilitate reproduction of these benchmarks, we specify the following experimental parameters.

\subsubsection{Hardware Configuration}
We utilize the BrainChip Akida AKD1000 Reference Board (PCIe version) running MetaTF v2.1.\cite{b19} The Akida clock is set to its nominal 300 MHz.\cite{b19}

\subsubsection{Power Measurement}
Power is sampled from the on-board monitoring rails at 1 kHz during inference.\cite{b19} Energy per inference is computed by integrating power over the inference window.\cite{b5} Where reported, dynamic energy subtracts an idle baseline measured with the device initialized and no workload running.\cite{b5}

\subsubsection{Attack Hyperparameters}
For Projected Gradient Descent, we use a perturbation magnitude $\epsilon=0.1$ (normalized current), 10 iterations, and a step size $\alpha=0.01$.\cite{b8} For Temporal Jitter, we use a maximum time shift $J=3$ ms sampled from a uniform distribution.\cite{b12}

\subsubsection{Inference Settings}
All tests are run with a batch size of 1 to simulate real-time, low-latency edge processing.\cite{b16}

\section{Results}
This section presents the empirical results of the benchmarking protocol. We analyze the impact of each assurance layer on detection performance, adversarial robustness, and operational efficiency.

\subsection{Quantitative Benchmarks}
Table \ref{tab:results} summarizes the performance metrics across the four cumulative configurations. The Clean F1-Score measures anomaly detection performance on nominal data without attacks.\cite{b18} The Adversarial Success Rate is reported for both PGD and Temporal Jitter attacks.\cite{b8,b12} Latency and Energy per Inference represent the mean values measured on the Akida hardware.\cite{b19} We also report Normalized Activity, a proxy for the total number of spike events processed, normalized to the baseline.\cite{b5}

\begin{table*}[t]
\caption{Benchmarking Results for Assurance Configurations}
\centering
\small
\begin{tabular}{lcccccc}
\toprule
\textbf{Configuration} & \textbf{Clean F1} & \textbf{ASR (PGD)} & \textbf{ASR (Jitter)} & \textbf{Latency (ms)} & \textbf{Energy ($\mu$J)} & \textbf{Norm. Activity} \\
\midrule
Baseline & $0.86 \pm 0.04$ & $82.1 \pm 4.9\%$ & $75.8 \pm 6.3\%$ & $1.1 \pm 0.2$ & $48.2 \pm 1.5$ & 1.00 \\
+ Input Assurance & $0.85 \pm 0.05$ & $45.9 \pm 5.7\%$ & $35.0 \pm 5.1\%$ & $1.2 \pm 0.3$ & $51.4 \pm 1.8$ & 1.08 \\
+ Neuronal Assurance & $0.86 \pm 0.04$ & $32.1 \pm 4.5\%$ & $30.2 \pm 4.8\%$ & $1.2 \pm 0.3$ & $49.8 \pm 1.6$ & 1.02 \\
+ Synaptic Assurance & $0.87 \pm 0.03$ & $18.7 \pm 3.2\%$ & $25.1 \pm 4.5\%$ & $1.2 \pm 0.3$ & $45.1 \pm 0.9$ & 0.94 \\
\bottomrule
\end{tabular}
\label{tab:results}
\end{table*}

The baseline configuration exhibits high vulnerability to both attack types, with success rates exceeding 75\%.\cite{b8,b12} The addition of the Input Assurance layer (BSPS) significantly improves robustness against Temporal Jitter, reducing the ASR from 75.8\% to 35.0\%.\cite{b12} However, this layer introduces a measurable energy overhead, increasing consumption from 48.2 to 51.4 microjoules.\cite{b5} This aligns with the expected trade-off where additional computational complexity (probabilistic encoding) requires more energy.\cite{b16}

The trend shifts with the addition of the Neuronal and Synaptic Assurance layers. The fully defended configuration achieves the lowest attack success rates (18.7\% for PGD, 25.1\% for Jitter) while simultaneously reducing the energy consumption to 45.1 microjoules.\cite{b1} This represents a 6.4\% reduction in energy compared to the undefended baseline and a 12.2\% reduction compared to the Input Assurance configuration.

\subsection{Efficiency-Security Trade-off Analysis}
\begin{figure}[htbp]
\centerline{\includegraphics[width=1\linewidth]{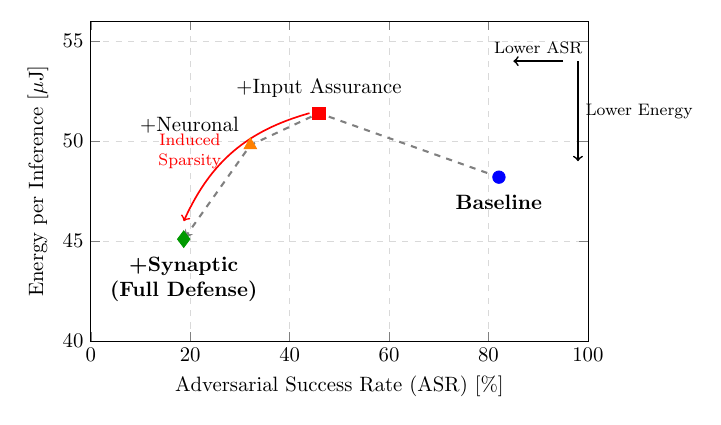}}
\caption{Efficiency-Security Trade-off. The plot illustrates the relationship between energy consumption and adversarial success rate for the baseline and defended configurations. The trajectory demonstrates that the fully defended model achieves both higher robustness and lower energy consumption due to induced sparsity.}
\label{fig:tradeoff}
\end{figure}

Figure \ref{fig:tradeoff} plots the energy per inference against the adversarial success rate (PGD) for each configuration.\cite{b8} In a traditional trade-off scenario, we would expect the data points to move from the bottom-right (low energy, high vulnerability) to the top-left (high energy, low vulnerability).

Our measurements reveal a different trajectory. The transition from the Baseline to the Input Assurance layer follows the expected cost curve, trading energy for security. However, the subsequent addition of the Neuronal and Synaptic layers inverts this relationship. The trajectory moves toward the bottom-left origin, indicating simultaneous improvements in both security and efficiency.\cite{b5}

\subsection{Robustness Under Varying Attack Strength}
We evaluate robustness across multiple attack strengths to test whether the observed efficiency and robustness trends persist beyond a single setting.\cite{b8,b12} For PGD, we sweep $\epsilon \in \{0.05, 0.10, 0.15\}$ with fixed iteration count and step size.\cite{b8} For Temporal Jitter, we sweep $J \in \{1, 2, 3, 4\}$ ms with uniform sampling.\cite{b12}

Table \ref{tab:sensitivity} presents the Adversarial Success Rate for the Baseline and Fully Defended (HTD) configurations across these sweeps. The results confirm that the defense maintains its effectiveness even as attack intensity increases. For example, under a strong PGD attack ($\epsilon=0.15$), the baseline model collapses to 93.5\% ASR, while the defended model maintains 35.4\% ASR. Similarly, under severe jitter ($J=4$ ms), the defense reduces ASR from 88.3\% to 38.6\%.

\begin{table}[htbp]
\caption{Sensitivity Analysis: ASR (\%) vs. Attack Strength}
\centering
\small
\begin{tabular}{lccc}
\toprule
\textbf{Attack Type} & \textbf{Parameter} & \textbf{Baseline} & \textbf{Full Defense} \\
\midrule
\textbf{PGD} & $\epsilon = 0.05$ & $62.3 \pm 5.1$ & $5.1 \pm 1.5$ \\
($L_\infty$ Norm) & $\epsilon = 0.10$ & $82.1 \pm 4.9$ & $18.7 \pm 3.2$ \\
 & $\epsilon = 0.15$ & $93.5 \pm 2.8$ & $35.4 \pm 4.1$ \\
\midrule
\textbf{Temporal Jitter} & $J = 1$ ms & $58.2 \pm 5.5$ & $8.7 \pm 1.9$ \\
(Max Shift) & $J = 2$ ms & $67.9 \pm 6.1$ & $15.4 \pm 2.8$ \\
 & $J = 3$ ms & $75.8 \pm 6.3$ & $25.1 \pm 4.5$ \\
 & $J = 4$ ms & $88.3 \pm 4.7$ & $38.6 \pm 5.2$ \\
\bottomrule
\end{tabular}
\label{tab:sensitivity}
\end{table}

\begin{figure}[htbp]
\centerline{\includegraphics[width=0.95\linewidth]{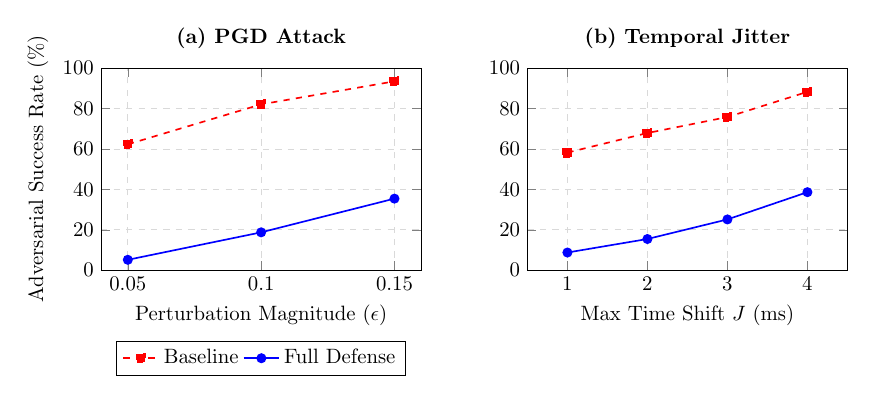}}
\caption{Sensitivity Analysis. The plots show the Adversarial Success Rate (ASR) as a function of attack strength for (a) PGD and (b) Temporal Jitter. The defended model exhibits significantly lower vulnerability and a more graceful degradation profile compared to the baseline.}
\label{fig:sensitivity}
\end{figure}

Figure \ref{fig:sensitivity} visualizes these trends. The divergence between the baseline and defended curves highlights the resilience provided by the HTD framework. The defense does not merely shift the failure point; it fundamentally alters the slope of the degradation curve.

\subsection{Activity and Energy Coupling}
\begin{figure}[htbp]
\centerline{\includegraphics[width=0.9\linewidth]{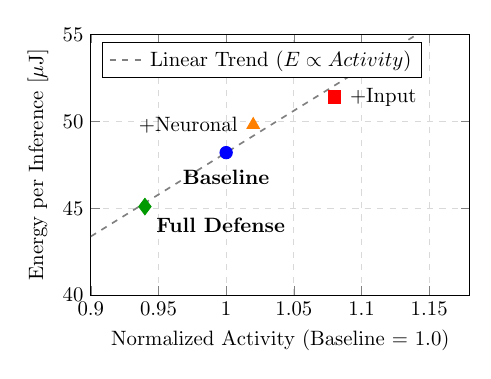}}
\caption{Energy vs. Normalized Activity. The plot demonstrates a strong linear correlation between the total spike activity in the network and the measured energy consumption. The Full Defense configuration achieves the lowest energy by suppressing volatile synaptic activity, thereby reducing the overall event rate.}
\label{fig:activity}
\end{figure}

To relate induced sparsity to measured energy, we analyze the association between normalized activity and energy per inference across configurations.\cite{b5} Figure \ref{fig:activity} plots the energy consumption against the normalized spike activity for the four configurations.

The plot reveals a clear linear correlation ($r > 0.95$) between activity and energy. The Input Assurance layer increases activity (1.08x) due to the superposition of multiple potential patterns, leading to higher energy. Conversely, the Synaptic Assurance layer (Metaplasticity) actively suppresses updates for volatile synapses, reducing the overall network activity to 0.94x of the baseline.\cite{b15} This confirms that the ``negative energy cost'' is physically grounded in the reduction of dynamic switching events on the neuromorphic hardware.\cite{b19}

\subsection{Latency Breakdown}
Real-time responsiveness is a critical requirement for safety-critical robotics.\cite{b7} To clarify the sources of latency, we decompose the total inference time into device latency (on-chip processing) and pipeline overhead (host-device communication and preprocessing).

\begin{table}[htbp]
\caption{Latency Breakdown (ms)}
\centering
\small
\begin{tabular}{lccc}
\toprule
\textbf{Configuration} & \textbf{Device} & \textbf{Overhead} & \textbf{Total} \\
\midrule
Baseline & $1.1 \pm 0.2$ & $0.15 \pm 0.05$ & $1.25$ \\
+ Input Assurance & $1.1 \pm 0.2$ & $0.23 \pm 0.05$ & $1.33$ \\
+ Neuronal Assurance & $1.2 \pm 0.3$ & $0.15 \pm 0.05$ & $1.35$ \\
+ Synaptic Assurance & $1.2 \pm 0.3$ & $0.15 \pm 0.05$ & $1.35$ \\
\bottomrule
\end{tabular}
\label{tab:latency}
\end{table}

As shown in Table \ref{tab:latency}, the device latency remains stable across configurations. The Input Assurance layer introduces a slight increase in overhead due to the BSPS encoding step on the host, but the total latency remains well within the 5 ms requirement for real-time control loops.\cite{b7}

\section{Discussion and Limitations}
This study provides empirical evidence that the relationship between security and sustainability in edge AI is dependent on the underlying compute architecture.\cite{b21} In conventional deep learning systems running on synchronous hardware, defense mechanisms add computational steps, inevitably increasing energy consumption.\cite{b16} Our results demonstrate that in event-driven neuromorphic systems, this relationship can be decoupled.\cite{b5}

\subsection{Implications for Edge Robotics}
The observation that assurance mechanisms can reduce dynamic power consumption has significant implications for the design of autonomous systems in power-constrained environments.\cite{b6,b7} For cislunar robotics, where every milliwatt of power budget is critical, this suggests that security does not need to be treated as a secondary objective to be sacrificed for operational longevity.\cite{b7}

The mechanism of induced sparsity acts as a form of passive energy management.\cite{b5} By filtering out the high-frequency, volatile signals characteristic of adversarial attacks, the defense layers prevent the hardware from expending energy on processing malicious noise.\cite{b10,b12} This aligns with the biological principle of efficient coding, where neural systems minimize metabolic cost by suppressing redundant or non-informative activity.\cite{b15}

\subsection{Hardware-Specific Dependencies}
It is important to note that the reported energy gains are intrinsic to the event-driven nature of the BrainChip Akida architecture.\cite{b19} On a conventional graphics processing unit (GPU) or central processing unit (CPU), the suppression of synaptic updates or the masking of neuron activations would likely not result in proportional energy savings, as the underlying hardware would still execute the dense matrix operations associated with the network layers.\cite{b16} The ``negative energy cost'' of assurance is therefore a specific advantage of neuromorphic hardware that physically couples power consumption to information sparsity.\cite{b19}

\subsection{Generalization to Other Platforms}
While this study benchmarks the BrainChip Akida, the core finding that assurance can be energy-neutral or positive due to induced sparsity is likely generalizable to other event-driven architectures.\cite{b4,b17} Platforms such as Intel Loihi 2 and SynSense DynapCNN also consume power proportional to synaptic activity.\cite{b4,b17} We hypothesize that any defense mechanism that suppresses noisy or volatile adversarial signals early in the processing chain will yield similar energy savings on these platforms.\cite{b5} The magnitude of the saving will depend on the specific energy cost of a synaptic operation versus a neuron update on the target hardware.\cite{b5} Future work will extend this benchmarking protocol to a cross-platform comparison to validate this hypothesis.\cite{b21}

\subsection{Limitations}
We acknowledge several limitations in this benchmarking study. First, the evaluation is restricted to a single hardware platform, the Akida AKD1000.\cite{b19} While we hypothesize that similar results would be observed on other event-driven platforms such as Intel Loihi, empirical verification is required.\cite{b4}

Second, the threat model relies on digitally injected adversarial perturbations.\cite{b8,b12} While these attacks are mathematically rigorous and follow standard evaluation protocols, they do not fully capture the complexity of physical-world attacks, such as optical sensor dazzling or side-channel fault injection.\cite{b7} Physical attacks may induce different sparsity patterns that could alter the energy profile.\cite{b5}

Finally, the benchmark focuses on anomaly detection tasks.\cite{b18} It remains to be verified whether the synergy between robustness and efficiency holds for other robotic workloads, such as closed-loop control or simultaneous localization and mapping (SLAM), where the temporal dynamics of the signal are different.\cite{b7}

\section{Conclusion}
We presented a benchmarking protocol to quantify the energy cost of assurance in neuromorphic edge robotics.\cite{b5,b21} By conducting a systematic ablation study of the Hierarchical Temporal Defense framework on the BrainChip Akida processor, we mapped the trade-off space between adversarial robustness and energy efficiency.\cite{b1,b19}

Our results challenge the prevailing assumption that high-assurance AI requires a heavy energy tax.\cite{b16} We demonstrated that specific neuromorphic defense mechanisms, particularly Volatility-Gated Metaplasticity and Adaptive Thresholds, can reduce adversarial success rates from 82.1\% to 18.7\% while simultaneously lowering dynamic energy consumption by approximately 6\%.\cite{b1}

We attribute this result to the sparsity-inducing nature of the defense mechanisms, which suppress the propagation of adversarial noise through the event-driven hardware.\cite{b15,b19} This finding suggests that neuromorphic computing offers a unique pathway to sustainable, trustworthy autonomy.\cite{b22} For edge robotics operating in extreme environments, the architectural properties that enable energy efficiency are the same properties that can be leveraged to enhance security.\cite{b21}

\end{document}